\documentclass[sigconf]{acmart}
\AtBeginDocument{%
  \providecommand\BibTeX{{%
    \normalfont B\kern-0.5em{\scshape i\kern-0.25em b}\kern-0.8em\TeX}}}

\copyrightyear{2024}
\acmYear{2024}
\setcopyright{acmlicensed}\acmConference[AVI 2024]{International Conference on Advanced Visual Interfaces 2024}{June 3--7, 2024}{Arenzano, Genoa, Italy}
\acmBooktitle{International Conference on Advanced Visual Interfaces 2024 (AVI 2024), June 3--7, 2024, Arenzano, Genoa, Italy}
\acmDOI{10.1145/3656650.3656675}
\acmISBN{979-8-4007-1764-2/24/06}
\begin{document}

\title{Gaze-Based Intention Recognition for Human-Robot Collaboration}

\author{Valerio Belcamino}
\orcid{1234-5678-9012}
\affiliation{%
  \institution{University of Genoa}
  \city{Genova}
  \country{Italy}
}
\email{valerio.belcamino@edu.unige.it}

\author{Miwa Takase}
\affiliation{%
  \institution{Shibaura Institute of Technology}
  \city{Tokyo}
  \country{Japan}}
\email{ma22088@shibaura-it.ac.jp}

\author{Mariya Kilina}
\affiliation{%
  \institution{University of Genoa}
  \city{Genova}
  \country{Italy}
}
\email{mariya.kilina@edu.unige.it}

\author{Alessandro Carfì}
\affiliation{%
  \institution{University of Genoa}
  \city{Genova}
  \country{Italy}
}
\email{alessandro.carfi@dibris.unige.it}

\author{Akira Shimada}
\affiliation{%
  \institution{Shibaura Institute of Technology}
  \city{Tokyo}
  \country{Japan}}
\email{ashimada@sic.shibaura-it.ac.jp}

\author{Sota Shimizu}
\affiliation{%
  \institution{Shibaura Institute of Technology}
  \city{Tokyo}
  \country{Japan}}
\email{sota@sic.shibaura-it.ac.jp}

\author{Fulvio Mastrogiovanni}
\affiliation{%
  \institution{University of Genoa}
  \city{Genova}
  \country{Italy}
}
\email{fulvio.mastrogiovanni@unige.it}

\renewcommand{\shortauthors}{Belcamino et al.}

\begin{abstract}
This work aims to tackle the intent recognition problem in Human-Robot Collaborative assembly scenarios. Precisely, we consider an interactive assembly of a wooden stool where the robot fetches the pieces in the correct order and the human builds the parts following the instruction manual. The intent recognition is limited to the idle state estimation and it is needed to ensure a better synchronization between the two agents. We carried out a comparison between two distinct solutions involving wearable sensors and eye tracking integrated into the perception pipeline of a flexible planning architecture based on Hierarchical Task Networks. At runtime, the wearable sensing module exploits the raw measurements from four 9-axis Inertial Measurement Units positioned on the wrists and hands of the user as an input for a Long Short-Term Memory Network. On the other hand, the eye tracking relies on a Head Mounted Display and Unreal Engine.

 We tested the effectiveness of the two approaches with 10 participants, each of whom explored both options in alternate order. We collected explicit metrics about the attractiveness and efficiency of the two techniques through User Experience Questionnaires as well as implicit criteria regarding the classification time and the overall assembly time.

The results of our work show that the two methods can reach comparable performances both in terms of effectiveness and user preference. Future development could aim at joining the two approaches two allow the recognition of more complex activities and to anticipate the user actions.

\end{abstract}




\maketitle

\begin{figure}[!hb]
  \includegraphics[width=0.85\columnwidth]{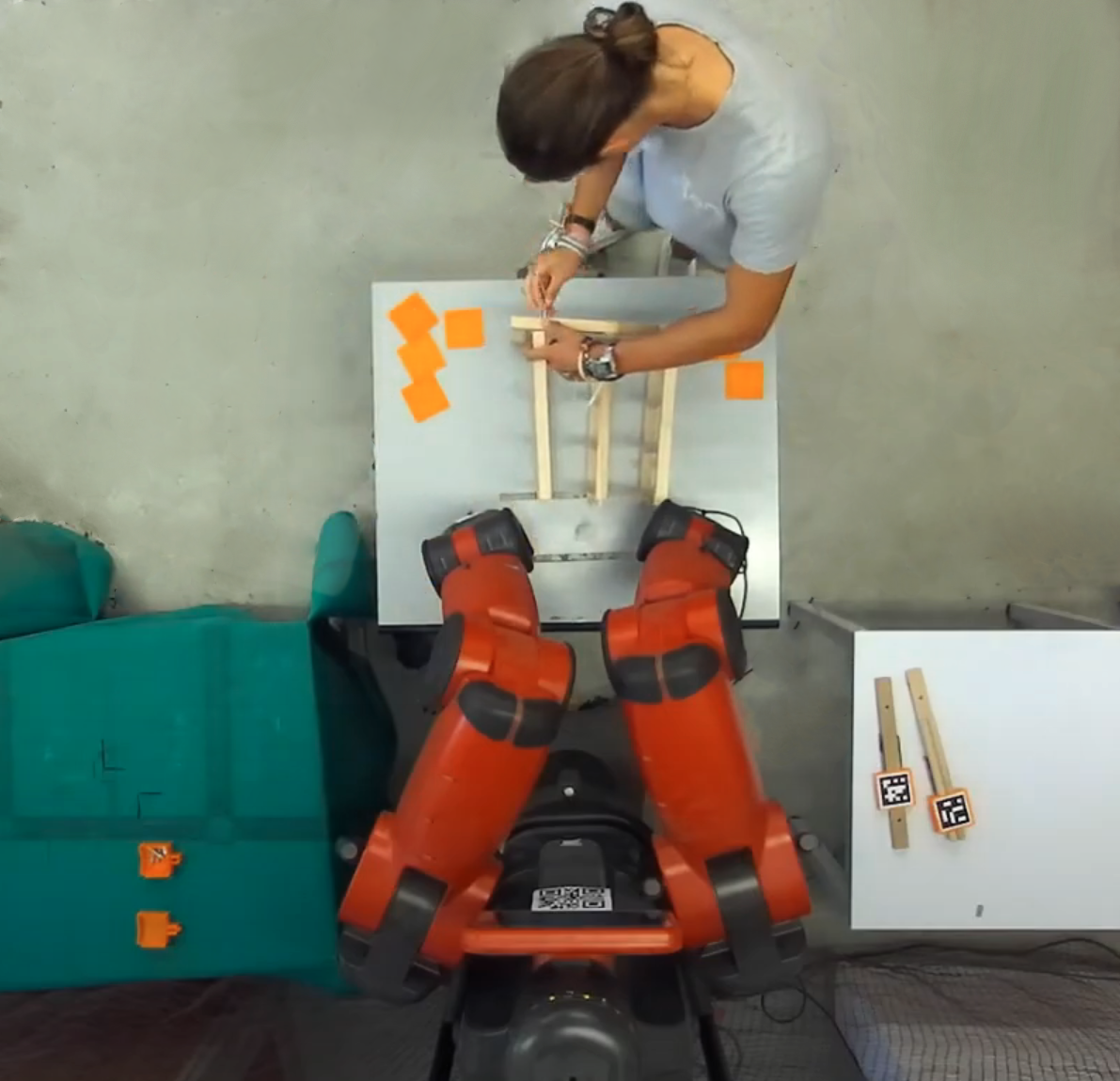}
  \caption{A top view of the experimental scenario, showing the interaction between human and robot. }
  \label{zed}
\end{figure}

\received{20 February 2007}
\received[revised]{12 March 2009}
\received[accepted]{5 June 2009}


\section{Introduction}
Technological progress is leading to a broader and more valuable integration of robots in the workplace. As a result, Human-Robot Interaction (HRI) rose as a central multidisciplinary domain encompassing aspects of engineering and cognitive sciences. This research field counts a multitude of applications comprising logistics \cite{logistics}, manufacturing \cite{manufacturing}, agriculture \cite{agricolture}, rehabilitation \cite{healthcare} and domestic care \cite{domesticassistance} and its main goal is to achieve seamless cooperation between the human agents and their robotic coworkers, exploiting the best capabilities of both worlds. 

In human-human collaboration, individuals tend to naturally decompose difficult problems into a list of simplified actions. The allocation of actions between agents and the timing of their execution is at the very core of efficient cooperation and is based on explicit communication and implicit cues. The explicit communication is mainly carried out through speech and gestures while the implicit cues derive from the observation of the surroundings and of the coworkers (i.e., detecting obstacles or noticing that a colleague is busy). Research in HRI aims at replicating these planning and perception capabilities to reach a level of coordination comparable to human-human interaction.

\begin{figure*}[!ht]
  \includegraphics[width=0.9\textwidth]{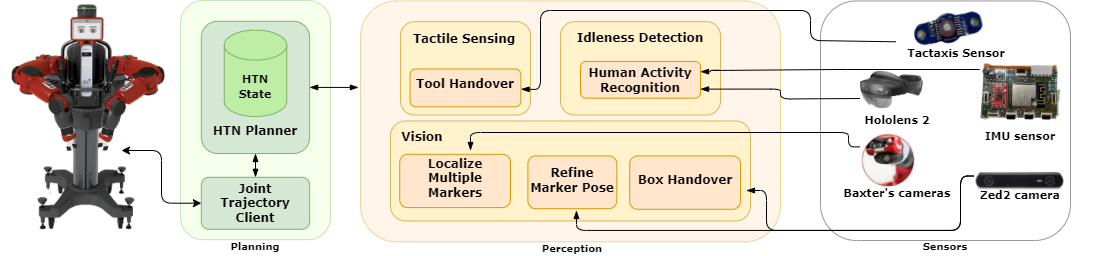}
  \caption{The architecture diagram of the system. The HTN planner updates its state activating the perception modules and can move the robot using the Joint Trajectory Client. The right side of the chart depicts the perception modules with the associated type of sensor.}
  \label{archi}
\end{figure*}

Various planning approaches have been explored in the literature, including Probabilistic planning \cite{probabilistic}, Knowledge-Based planning \cite{knowledge}, Learning-Based planning \cite{reinforcement}, and Hierarchical planning \cite{htn}. Probabilistic planning allows robots to adapt to uncertain information but faces challenges in interpretability and computational resources. Knowledge-based planning incorporates semantic knowledge for autonomous problem-solving, but scalability may be hindered by potential conflicts. Learning-based planning models human behaviours using techniques like reinforcement learning but is sensitive to data quality and lacks interpretability. Hierarchical planning, mirroring human problem-solving, breaks down complex tasks into subtasks, enhancing efficiency, but the hierarchical nature introduces complexities in decision-making processes.

To effectively schedule and assign actions between the users, the planner needs to be able to detect changes in the state of the environment and track the development of the actions carried out by the agents. The perception pipeline is part of the architecture that observes the development of the plan and updates the internal state variables.
The perception problem has been tackled in the literature with several methods depending on the information that the robot needs to extrapolate from its surroundings. We chose to focus on human intent recognition as it is one of the bottlenecks that hinder effective collaboration. The recognition of the intent of the human agent can allow for better synchronization and lead to a safer and more robust interaction. To recognize the intent of the human it is important to track its movements online. In the literature, the main motion-tracking approaches rely on vision and wearable sensors. The first category provides images and enables object recognition, scene understanding, and navigation \cite{navigation}. 
In the second case, we obtain different information depending on the type of sensor as we can model tactile perception \cite{theo} through touch sensors, perform body tracking with Inertial Measurement Units (IMU) or achieve gaze estimation \cite{gaze} using eye trackers.

Gaze is one of the most effective ways for humans to sense the intentions of their partners in a non-verbal manner and we focused on this aspect of perception to tackle human intent recognition. 
Humans are predominantly visual creatures \cite{Balaram} who rely heavily on their eyes to perform most of their actions \cite{okumura2013power} and to convey their emotions \cite{liang2021emotional}. Furthermore, when people are interacting with their surroundings or other individuals, their gaze frequently anticipate their movements \cite{Barnes} and as a result of this characteristic, eye tracking might be employed to foresee their intents. 
Precisely for this reasons, gaze estimation is a rapidly progressing area of research and finds application in a variety of fields: from the study of social interactions \cite{canigueral2019role} to robotics \cite{belardinelli2023gaze,palinko2015eye}. Furthermore, in robotics, the importance of the gaze is twofold as it allows both the prediction of user actions \cite{7822895,shi2021gazeemd,singh2020combining} and the reproduction of credible behaviour on humanoid robots \cite{admoni2017social,Friebe_2022}. In our case, we exploited the eye-tracking capabilities provided by a Head Mounted Display for Augmented Reality to obtain a smoother HRI by estimating the intent of the human agent.
The activity recognition module based on gaze was integrated in the perception pipeline of a modular software architecture for HRC based on Hierarchical Task Networks (HTN). Then, we carried out an experimental phase and compared the performance of the gaze estimation to the IMU classifier, which is the most common wearable solution for Human Activity Recognition (HAR) \cite{Lastrico,RUZZON2020106122}.

\section{Methodology}
The collaborative task that we are targeting is the assembly of an IKEA stool. The task requires the pick-and-place of the components and the use of tools and screws to secure the wooden parts. The end-effector mounted on the robot lacks the dexterity to manipulate small objects and use tools, for this reason, the role of the robot in the collaboration is to locate, grasp and deliver the objects located in the workspace. In contrast, the human waits for the robot and assembles the pieces in the correct order.
To carry out the task coherently and effectively, the robot needs to synchronize its actions according to human needs and for this reason, intent recognition plays and central role in the scenario.

We formalize the Human Robot Collaboration problem focusing on the state of the interaction and the task plan. The state includes the properties required to describe the progress of the interaction and the effects of agents' actions on the workspace. We identified four features for the state, which are end-effector availability (EEA), agent pose (AP), object pose (OP), and object characteristics (OC). As for the task plan, we modelled the scenario with six primitive actions: \textit{grasp}, \textit{release}, \textit{move}, \textit{manipulate}, \textit{wait} and \textit{perceive}. The first three can be combined to describe pick-and-place and handover, while the fourth refers to motions performed by the end-effector that alter the state of an object (e.g., using a screwdriver). Finally, \textit{perceive} is the action that allows the agents to update their internal representation of the interaction state and is often used in combination with \textit{wait} to synchronize cooperative actions. 

This work focuses on the human intent recognition and aims at improving the synchronisation between human and robot. Moreover, we assume a correlation between the intent of the user and the action that they are currently completing. Specifically, if the person is assembling the parts already provided by the robot, we assume that the robot needs to wait for the completion of this task. Vice versa, if the user is currently idle, they are probably waiting for the handover of the next parts and the robot should resume its plan. We use two different approaches for the recognition of the person's actions: one based on gaze and one on inertial sensors. In the first case, we performed gaze estimation to classify the idle state of the user, while in the second we chose IMU and deep learning. Gaze estimation can be particularly helpful in this classification task since the person will be keen to focus on the workspace during the execution of the assembly tasks and will try to check their coworker and the remaining pieces during the idle time to plan the subsequent actions \cite{lego}. We decided to compare this system against IMUs because they are the most widespread wearable sensors employed in HAR.

\begin{figure*}[!ht]
  \includegraphics[width=0.9\textwidth]{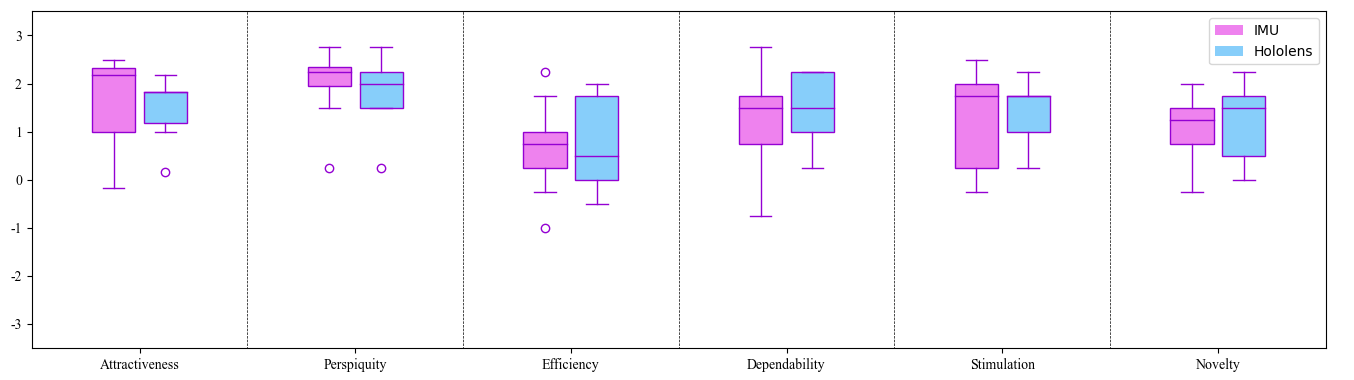}
  \caption{The picture shows a comparison of the results obtained by the idleness classification based on IMU sensors and Hololens. Each subplot from left to right is associated to one of the criteria provided by the User Experience Questionnaire.}
  \label{ueq}
\end{figure*}

\section{Experimental Setup}
The experimental scenario that we developed consists of the collaborative assembly of a stool, in which the role of the robot is to locate and bring the required objects to the user, who must connect them following instructions manual\footnote{\scriptsize\href{https://www.ikea.com/it/it/p/oddvar-sgabello-pino-20249330/}{ikea.com/it/it/p/oddvar-sgabello-pino-20249330}}.

The developed architecture is displayed in Figure \ref{archi} and includes a collaborative Baxter robot by Rethink Robotics surrounded by three tables constituting the workspace shared with the human as shown in Figure \ref{zed}. The perception pipeline is composed of a vision and a tactile sensing module. The first one encompasses a Zed2 RGB-D camera and the two wrist-mounted cameras of the robot, while the second relies on the Tactaxis sensor \cite{theo} by Melexis\footnote{\scriptsize\href{https://www.melexis.com/en}{melexis.com/en}}  attached to the robot's gripper. Additionally, we integrated 4 9-axis IMUs \cite{carfì2024modular} and a Hololens Head-Mounted Display (HMD) by Microsoft for the human intent estimation.

At runtime, the Zed2 camera framed the workspace from the ceiling as shown in Figure \ref{zed} and tracked object poses using AruCo markers. This pose was then refined using the wrist cameras whenever the robot approached the objects. Moreover, wrist cameras have been used to perform colour segmentation and estimating the content of small boxes filled with screws. The tactile sensor allowed to enhance the handover of the tools by monitoring the sheer forces and releasing the grasp with the appropriate timing. 
Lastly, there are the two intent recognition modules. In the case of the IMU sensors, the stream of raw data is fed to a Long Short-Term Memory (LSTM) Network implemented in Pytorch, which performs binary classification. The dataset used to train the LSTM was collected in one of our previous works involving the same sensors and a similar setup \cite{simone} and counts 6462 sequences of 500 samples acquired at a frequency of 30 Hz. Moreover, this classifier reaches a precision of 96.6\%, a recall of 99.6\% and an F-Score of 0.98. As for the Hololens, we developed an Unreal Engine application which overlays the real workspace with its holographic representation and integrated the Mixed Reality Toolkit (MRTK)\footnote{\scriptsize\href{https://learn.microsoft.com/en-us/windows/mixed-reality/develop/unreal/unreal-mrtk-introduction}{learn.microsoft.com/en-us/windows/mixed-reality/develop/unreal/unreal-mrtk-introduction}} to perform online gaze estimation. 
At runtime, the application can track the human gaze and trace a ray cast in the same direction. A collision query is performed for each ray; in case of collision between the ray and the holograms representing the workspace, we update the action performed by the user. We consider the user to be working whenever they are focusing their attention on the workspace, while we consider them idle when they start exploring their collaborator's working area. We chose this approach since we observed that the assembly actions occurring in the task needed eye-hand coordination and thus, the gaze of the user must be focused on the table most of the time. Vice versa, when the person completes their task they are more keen to check the neighbouring tables and the robot to choose the most suitable subsequent action. This simplification seems to be substantial for our current setup because there is a limited number of action that need direct coordination between the two agents; however, a more complete approach could consider more complex phenomena like preparatory gaze to convey the human's future intents.

\begin{figure}[!hb]
  \includegraphics[width=\columnwidth]{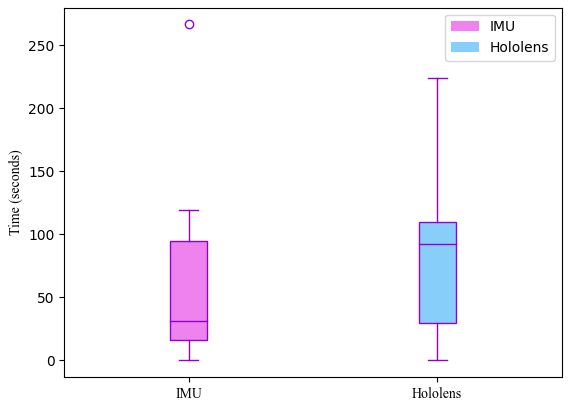}
  \caption{The two boxplots show the time needed to complete the Wait action for the two interfaces}
  \label{time}
\end{figure}

\section{Results}
We conducted our comparison relying on the architecture described in the previous section using the two distinct intent recognition modules. We planned a within-subject experimental design to evaluate the two interfaces where each of the ten participants tried both options. We chose this design to reduce variability between the participants and because it typically requires smaller sample sizes to achieve significant statistical power. Additionally, we mitigated the bias effect due to order by asking 5 participants to start with the IMU sensors and the other 5 with the Hololens.
%Ten subjects participated in the tests, and each person completed the task twice to test both approaches. Additionally, we asked 5 participants to start with the IMU sensors and the other 5 with the Hololens to avoid biases.
At the end of each trial, the participants were asked to fill out the User Experience Questionnaire (UEQ) \cite{ueq}, which is a common survey employed in the assessment of the User Experience (UX) of interactive products and allowed us to compare the two approaches considering six different criteria: attractiveness, perspicuity, efficiency, dependability, stimulation and novelty. The questionnaire requires answering 26 questions with an integer score in the range [-3, +3] where the lowest number represents an extremely negative opinion and the highest an extremely positive one. The answers' results are shown in Figure \ref{ueq}, where each subplot is associated with one of the six categories and the two methods are displayed in different colours. The two approaches reach a similar score for each feature, particularly the average values are extremely close. This behaviour indicates that subjects found the two interaction modes comparable and that there was no preference in terms of user experience. 

The second type of data that we gathered from the experiments is objective and refers to the idle time and the duration of the full assembly. With the term idle time we are considering the amount of time that the robot passed in the waiting state while the human was performing an assembly action and the results are shown in Figure \ref{time}. We can notice how the two distributions are quite close, having an average of 80 seconds (SD=64.3) for the IMUs and 83 seconds (SD=61.4) for the Hololens. The spread of the values shown in the boxplots is mainly because the \textit{wait} action occurs when the user needs to accomplish assembly tasks that have variable lengths depending on the number of pieces. As for the global assembly time, the IMU reaches a value equal to 618 seconds (SD=159.9), while the Hololens have a higher duration of 806 seconds (SD=263.0). This difference is due to the cumulative sum of the time of the idle time, which in the case of the Hololens has a higher variance.

\section{Conclusions}
We presented a perception module for human intent recognition based on gaze, we integrated it into a modular planning architecture and tested its performance in a collaborative assembly scenario against a similar approach based on IMUs.
The result of the comparison indicates that the gaze estimation achieves IMU-like performance both in terms of user preference and classification time. We can also highlight how the approach based on IMUs requires more sensors as it needs to track both hands while the gaze estimation only requires an HMD. Moreover, most commercial HMDs include other extremely useful features that could be integrated into HRI tasks; for example, in the case of the Hololens2, the MRTK allows to easily implement vocal commands, hand tracking, gesture recognition as well as its core augmented reality capabilities. On the other hand, the IMU module relies on LSTM, which can also be used to forecast the subsequent actions enabling the robot to anticipate human behaviour. In comparing the IMU and gaze methods using eye tracking, it's also important to consider factors such as cost and processing power, with the IMU generally being more cost-effective. Generally, we could consider gaze estimation as an alternative approach to intent recognition, while on the other side, it could represent a valuable enhancement of the most common approaches based on IMUs. In fact, the integration of the two techniques could help to model the detection of more complex intents and predict more accurately the human actions.

\section*{Acknowledgment}
This work received funding from the Italian Ministry of Education and Research (MIUR). This work has been also made with the Italian government support under the National Recovery and Resilience Plan (NRRP), Mission 4, Component 2 Investment 1.5, funded from the European Union NextGenerationEU. Additional fundings came from the TOBITATE scholarship granted by the Japanese government. Lastly, the authors would like to acknowledge Melexis\footnotemark[2] for providing the Tactaxis sensor, which constitutes the tactile perception module of our architecture.
\bibliographystyle{ACM-Reference-Format}
\bibliography{sample-base}


\end{document}